%% file: root.tex
\pdfoutput=1

\documentclass[letterpaper, 10 pt, conference]{ieeeconf}  % Comment this line out if you need a4paper

\IEEEoverridecommandlockouts                              % This command is only needed if 
\usepackage[T1]{fontenc} % Preserve the class's Times font with both pdfLaTeX and XeTeX.

\usepackage{xcolor}
\usepackage{graphicx}
\usepackage{svg}

\usepackage{tabularx} 
\usepackage{listings}
\usepackage{url}
\usepackage[ruled,vlined]{algorithm2e}

\usepackage[breakable]{tcolorbox}
\usepackage[caption=false,font=footnotesize]{subfig} 
\usepackage{amsmath}
\usepackage{cite}

\usepackage{multirow}
\tcbuselibrary{listings, breakable}

\usepackage{pifont}

\newtcolorbox{userbox}{
  colback=gray!3!white,   
  colframe=gray!40!black,  
  title=Prompt Template,
  fonttitle=\bfseries,
  breakable
}

\lstdefinelanguage{PDDL}{
  morekeywords={define,domain,requirements,types,predicates,action,parameters,
                precondition,effect,init,goal,problem,when,not,and,or},
  sensitive=false
}

\lstdefinestyle{lispstyle}{
  language=PDDL,
  basicstyle=\ttfamily\small,      
  keywordstyle=\bfseries\color{blue!60!black},
  commentstyle=\itshape\color{gray!80!black},
  columns=fullflexible,
  keepspaces=true,
  showstringspaces=false,
  frame=single,                     
  rulecolor=\color{black!20},
  backgroundcolor=\color{gray!3},
  frameround=tttt,                  
  breaklines=true,                  
  captionpos=b,                     
  xleftmargin=1em, xrightmargin=1em
}

\title{\LARGE \bf
SafeInferCom: Safe Inference-Time Compute via Verifier-Guided Mid-Generation Intervention for Robotic Task Planning
}

\author{Weizhe Xu$^{1}$, Jialiang Fan$^{1}$, Mengyu Liu$^{2}$, and Fanxin Kong$^{1}$\\
\textit{$^{1}$University of Notre Dame \quad $^{2}$Washington State University}\\
\{wxu3, jfan5, fkong\}@nd.edu, mengyu.liu@wsu.edu
}

\begin{document}

\maketitle
\thispagestyle{empty}
\pagestyle{empty}

\vspace{-2 em}

\input{sections/0_abstract}
\input{sections/1_introduction}
\input{sections/2_related_work}
\input{sections/3_preliminaries}

\input{sections/4_inference_time_reasoning_monitor}

\input{sections/5_safeInferCom}
\input{sections/6_experiment}
\input{sections/7_discussion}
\input{sections/7_conclusion}

\bibliographystyle{IEEEtran}
\bibliography{ref}

\end{document}

%% file: sections/0_abstract.tex
\begin{abstract}
Large Reasoning Language Models (LRLMs) enable multi-step reasoning for robotic task planning, but continued reasoning can overwrite valid intermediate plans or leave constraint violations unresolved, reducing planning reliability and wasting inference-time computation.
We develop an inference-time monitor that exposes and verifies intermediate plans without disrupting the original decoding trajectory.
Building on this monitor, we propose SafeInferCom, a formal verifier-guided framework that preserves valid intermediate plans and directs error correction during generation.
Experiments across multiple LRLMs and planning domains reveal reasoning--response inconsistency and limited self-correction under one-shot inference.
SafeInferCom improves planning success and accelerates error correction relative to one-shot inference.
When combined with iterative refinement, it further improves success while reducing token usage compared with refinement alone.
We additionally evaluate SafeInferCom in VirtualHome and provide a real-world robotic-arm demonstration.
\end{abstract}

%% file: sections/1_introduction.tex
\section{Introduction}
Robotic task planning converts high-level goals into executable action sequences while ensuring that safety constraints are satisfied throughout execution.
Classical planners provide a principled and auditable solution when an accurate symbolic model is available: planners such as Fast Downward search over explicit action models, while validators such as VAL verify candidate plans against the given specification~\cite{helmert2006fast,howey2004val}.
However, this formalism has two practical limitations.
First, it requires rigid symbolic task representations and carefully engineered domain models. 
Second, symbolic search can become computationally expensive as the number of objects, applicable actions, and planning horizon increase~\cite{kwon2025fast}.
Learning-based planners enable fast inference through learned policies, but depend on training data and generally lack explicit safety guarantees~\cite{dong2023optimizing,janner2021offline}.

Large Language Models (LLMs) have been explored for robotic task planning, leveraging their ability to interpret high-level instructions and contextual descriptions to generate candidate action sequences~\cite{liu2023llm+,xu2024assuring}. 
While conventional LLM planners typically generate a final plan directly from the input, Large Reasoning Language Models (LRLMs) first perform explicit multi-step reasoning before producing the final output~\cite{openai2024reason,guo2025deepseek}. 
By allocating additional computation at inference time to this intermediate reasoning process, LRLMs can deliberate over candidate actions and constraints before committing to a plan.

However, additional reasoning does not always lead to a better plan. 
Through fine-grained analysis, we observe reasoning--response inconsistency across models and planning domains. 
In some cases, a model develops a safe and goal-satisfying intermediate plan but later overwrites it through continued reasoning, ultimately producing an incorrect final output. 
Conversely, constraint-violating intermediate plans may persist across successive reasoning steps without effective correction. 
Such instability reduces planning reliability and wastes inference-time computation on trajectories that could have been terminated or corrected earlier.

Existing approaches improve LLM-based planning reliability by using external formal verifiers to validate completed plans, followed by iterative refinement or verification-based candidate selection~\cite{jha2023neuro,xu2024assuring,zhou2024isr,hu2024deploying}. 
While these methods can improve final-plan validity, the verifier typically provides feedback only after a candidate has been fully generated. 
For LRLMs, a valid intermediate plan may therefore be overwritten, or a constraint violation may persist, before external verification occurs. 
This motivates integrating formal verification into the reasoning process to preserve valid intermediate plans and guide correction before generation is complete.

To systematically analyze reasoning-level safety issues, we first develop an inference-time reasoning monitor that instruments LRLM generation without altering its primary decoding trajectory. 
The monitor extracts structured action sequences at selected reasoning transitions and evaluates them with an external formal verifier, while allowing the model to resume reasoning from the original state. 
This enables fine-grained analysis of reasoning--response consistency under standard one-shot inference.

Building on this monitoring infrastructure, we introduce \textbf{SafeInferCom}, a verifier-guided mid-generation controller with four actions: continue a safe but incomplete trajectory, inject diagnostic feedback after a detected violation, terminate generation once a valid plan is identified, or trigger replanning after persistent violations.
Unlike post-hoc refinement, SafeInferCom intervenes before a complete failed trajectory is generated, allowing violations to be corrected during reasoning and earlier valid candidates to be preserved without restarting the entire inference process. 
In this way, SafeInferCom shifts formal verification from post-hoc plan checking toward mid-generation guidance, improving planning reliability while avoiding unnecessary continued reasoning once a valid intermediate plan is identified.

Our contributions are summarized as follows: 
(1) An inference-time reasoning monitor that extracts and formally verifies intermediate plans from LRLM reasoning prefixes while preserving the original decoding trajectory, enabling fine-grained analysis of reasoning--response inconsistency.
(2) \textbf{SafeInferCom}, a verifier-guided mid-generation controller with four actions: continuation, targeted correction, early termination, and replanning after persistent violations.
(3) A comprehensive evaluation across multiple LRLMs and robotic planning domains, characterizing intermediate-plan fidelity, self-correction behavior, and the success--compute trade-off against one-shot inference and post-hoc refinement. Additional VirtualHome and real-robot manipulation studies demonstrate the method in more realistic settings.

%% file: sections/2_related_work.tex
\section{Related Work}~\label{related}

\subsection{Classical, Learned, and Language-Based Task Planning}

Classical planners search over explicit symbolic transition models, providing formal plan validation~\cite{helmert2006fast,howey2004val}.
However, they require carefully constructed symbolic representations and can incur explosive search cost as the number of objects, applicable actions, or planning horizon grows~\cite{kwon2025fast}.
Reinforcement and imitation learning instead learn policies or search guidance from demonstrations or environment interaction~\cite{dong2023optimizing,janner2021offline}. 
These methods amortize planning computation through learned policies, but their performance remains tied to the training distribution and formal guarantees require an external verifier.
Language-based planners offer greater input flexibility by generating action sequences from natural-language instructions and contextual descriptions~\cite{jansen2020visually,brohan2023rt}, motivating recent efforts to combine language-model planning with symbolic reasoning and verification.

\subsection{Language Models with Symbolic Planning and Verification}

Language models can play different roles in symbolically grounded planning systems.
Some works translate tasks into formal representations and delegate planning to symbolic solvers~\cite{liu2023llm+,jiang2026language}.
ISR-LLM instead uses an LLM to generate plans from symbolic domain and problem descriptions and refines them using formal validator feedback~\cite{zhou2024isr}. 
Following a similar problem setting, other approaches also use external formal verification or temporal-constraint-based checking to validate LLM-generated plans against explicit task specifications~\cite{obi2025safeplan,jha2023neuro,xu2024assuring,yang2024plug}.

These methods typically invoke formal verification after a complete candidate plan has been generated, or between successive generation rounds. 
Consequently, they do not directly regulate the intermediate reasoning trajectory within a single inference pass. 
For CoT and LRLM-based planners, continued reasoning may therefore overwrite an already valid intermediate plan or allow an invalid candidate to persist before verification occurs. 
SafeInferCom instead applies formal verification to candidate plans induced from intermediate reasoning prefixes and uses the resulting feedback for mid-generation control.

Following this line of work~\cite{zhou2024isr,obi2025safeplan,jha2023neuro,xu2024assuring,yang2024plug}, we study language-model-based planning under the same general setting, where candidate plans can be checked against an explicit task specification using an external formal verifier.

\subsection{Evaluation of Language-Model Reasoning}

Beyond outcome-level metrics such as accuracy and Pass@k, prior work has examined the quality of intermediate reasoning in language models. 
Some evaluation frameworks~\cite{golovneva2022roscoe,prasad2023receval} assess properties including consistency, correctness, and informativeness within reasoning traces. 
Step-aware verifiers further evaluate intermediate reasoning steps to support final-answer selection~\cite{li2023making,zhong2026verytrace}, while other studies investigate critique, error detection, reflection, and self-correction capabilities across linguistic, mathematical, and symbolic reasoning tasks~\cite{lin2024criticbench,chen2025finereason}.

These efforts primarily evaluate reasoning quality at the language or logical level rather than the validity of the action sequences implied by intermediate reasoning under an explicit robotic task specification. 
A reasoning trace may therefore appear coherent while inducing a candidate plan that violates action preconditions, safety constraints, or task goals. 
Moreover, intermediate-reasoning evaluation alone does not provide an online mechanism for correcting an ongoing trajectory or preserving an already valid candidate during generation.

\subsection{Inference-Time Compute Scaling and Regulation}

Recent work has shown that additional inference-time computation can improve the performance of reasoning models~\cite{balachandran2025inference}. 
Inference-time scaling methods allocate more reasoning steps or sample additional trajectories to improve solution quality on challenging tasks. 
Adaptive approaches further regulate computation during generation through restarting, pruning, or early termination~\cite{manvi2024adaptive,yang2026dynamic,bhat2026interwhen}. 

These methods primarily regulate inference based on predicted response quality, confidence, or the expected utility of additional computation. 
In contrast, SafeInferCom uses external formal verification of intermediate candidate plans as the control signal. 
This enables both corrective intervention and verified early termination, directly linking inference-time compute regulation to planning validity under an explicit task specification, while reusing the prefix decoding state to reduce monitoring overhead.

%% file: sections/3_preliminaries.tex
\section{PRELIMINARIES AND PROBLEM STATEMENT}~\label{preliminaries}
\begin{figure*}[t] 
    \centering
    % \includesvg[width=0.8\textwidth]{figures/main19} 
    \includegraphics[width=\linewidth]{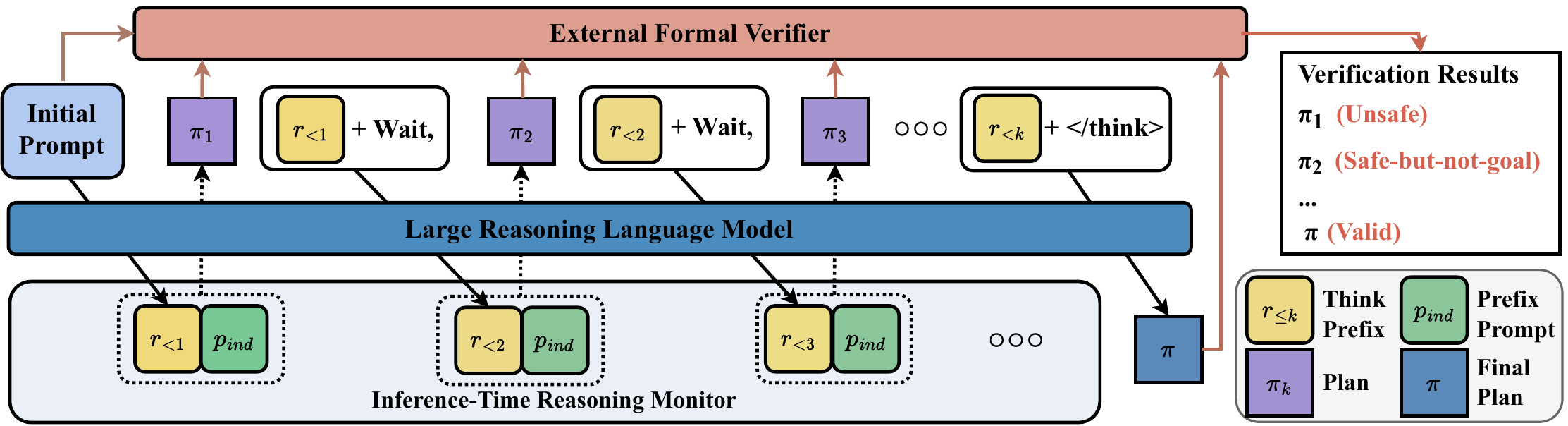}
    \vspace{-0.5em}
    \caption{
Overall workflow of the inference-time monitor. 
At a selected checkpoint, primary decoding is paused and an auxiliary branch appends the induction prompt $p_{\mathrm{ind}}$ to the reasoning prefix $\mathbf{r}_{< k}$ to produce an induced candidate plan $\pi_k$. 
The external verifier returns its label $y_k$, after which the auxiliary branch is discarded and primary decoding resumes until the final plan $\pi$ is produced.
    }
    \vspace{-1.5em}
    \label{fig:architecture}
\end{figure*}

\subsection{Robotic Task Planning with Safety Constraints}

We consider deterministic and fully observable robotic planning problems defined by a planning domain $\mathcal{D}$, an initial state $s_0$, a goal condition $G$, and optional explicit safety constraints $\mathcal{S}$.
A plan $\pi=\langle a_1,\ldots,a_T\rangle$ induces a state trajectory $\tau(\pi)=\langle s_0,s_1,\ldots,s_T\rangle$ according to the transition model specified by $\mathcal{D}$.
A valid plan must be executable under the action semantics of $\mathcal{D}$, satisfy the safety constraints in $\mathcal{S}$ throughout execution, and reach a final state satisfying $G$.

We assume access to an external formal verifier $V$ that evaluates a candidate plan with respect to
$\mathcal{P}=(\mathcal{D},s_0,G,\mathcal{S})$ and returns one of three outcomes:
\textsc{Unsafe}, if the plan violates an action requirement or an explicit safety constraint; in our standard PDDL benchmarks, \textsc{Unsafe} primarily corresponds to action-precondition or applicability violations.
\textsc{Safe-but-Not-Goal}, if the plan is executable and safe but does not satisfy $G$; and
\textsc{Valid}, if the plan is both safe and goal-achieving.
For an invalid candidate, $V$ also returns the first failing step and corresponding diagnostic information.
In our implementation, $V$ is instantiated using a standard PDDL plan validator~\cite{howey2004val}.

Any validity guarantee in this work is relative to the supplied task specification $\mathcal{P}$ and assumes that $V$ correctly checks candidate plans against it.
It does not cover task-grounding errors, omissions in the symbolic model, or failures in perception and physical execution.

\subsection{LRLM Inference-Time Reasoning Setting}

Given a robotic planning problem $\mathcal{P}$ and its prompted description $x(\mathcal{P})$, an LRLM generates an intermediate reasoning trace followed by a final plan:
$x(\mathcal{P}) \rightarrow \mathbf{r} \rightarrow \pi$,
where $\mathbf{r}=\langle r_1,\ldots,r_K\rangle$ is the sequence
of reasoning tokens, $K$ is the number of reasoning tokens generated
during inference, and $\pi$ is the final generated plan.
Unlike conventional LLM planners that directly produce a final response, LRLMs explicitly expose an intermediate reasoning process that shapes the resulting plan. Although this reasoning is not executed in the environment, it reflects the model's evolving planning decisions before producing the final output.

We use \emph{reasoning--response inconsistency} to describe cases in which the planning decisions expressed during intermediate reasoning disagree with the validity of the final plan. In particular, an LRLM may identify a valid plan during reasoning but lose it through continued deliberation, or may reach a valid final plan without exhibiting any valid intermediate plan beforehand. The former can reduce planning reliability, while the latter makes the decision process less interpretable and reduces user confidence in the final result.

\subsection{Problem Statement}

We consider LRLM-based robotic planning settings in which candidate plans can be evaluated by an external formal verifier against an explicit task specification. Under standard one-shot inference, however, the LRLM completes its reasoning trajectory and final response without using this verifier during generation.

We therefore study two related objectives. 
The \emph{monitoring objective} is to evaluate the natural-language reasoning process exposed during LRLM inference in safety-critical robotic planning, without altering the standard one-shot inference trajectory.
The \emph{control objective} is to use intermediate verification feedback to guide generation toward valid plans while avoiding unnecessary continued reasoning after a valid candidate has been identified.

%% file: sections/4_inference_time_reasoning_monitor.tex
\section{Inference-Time Reasoning Monitor}
\label{sec:monitor}

The inference-time reasoning monitor exposes intermediate planning states during LRLM generation by converting selected reasoning prefixes into candidate plans that can be evaluated by the external formal verifier (Fig.~\ref{fig:architecture}). It operates non-invasively under standard one-shot inference, enabling analysis of how planning decisions evolve during reasoning while also providing the verifier outcomes later used by SafeInferCom for mid-generation control.

\subsection{Reasoning Transition Monitoring}
We monitor reasoning prefixes $\mathbf{r}_{< k}$ at selected checkpoints defined by reasoning transitions.
Reasoning models often naturally express reflection, verification, or changes of direction through linguistic markers such as ``wait'', ``alternatively'', and ``hmm''. 
Recent work similarly uses such markers as lightweight boundaries for segmenting reasoning traces into reasoning actions or chunks~\cite{yang2026dynamic,jiang2025makes}. 
We therefore use these transitions to identify points at which the model's current planning decisions are examined.
These transitions are not explicitly requested in the prompt. 

Let $\mathcal{W}$ denote a finite set of transition-marker token sequences and $R_{\max}$ the maximum number of monitoring checkpoints. 
A checkpoint is triggered when the newly generated suffix matches a marker in $\mathcal{W}$, subject to the budget $R_{\max}$. 
At each checkpoint, primary decoding is paused and its current decoding state, including the KV cache and random number generator (RNG) state, is saved. 
The detected transition marker defines the boundary between the preceding reasoning prefix $\mathbf{r}_{<k}$ and the next reasoning stage.

\subsection{Intermediate Plan Induction}

Reasoning prefixes are expressed in natural language and therefore cannot be submitted directly to the formal verifier. 
To obtain a verifiable representation, the monitor creates an auxiliary decoding branch from the saved checkpoint state. 
On this auxiliary branch, the detected transition marker is removed and a fixed induction prompt $p_{\mathrm{ind}}$ is appended to the preceding reasoning prefix $\mathbf{r}_{<k}$. 
The same LRLM is then instructed to express the action sequence implied by its reasoning so far in the plan format expected by the verifier.

The induction process is intended to expose, rather than repair or complete, the planning decisions already reflected in the reasoning prefix. 
It preserves planning errors, incorporates revisions already expressed in the reasoning, and excludes explicitly rejected alternatives. 
Formally,
\[
\pi_k \sim \mathcal{M}\!\left([x(\mathcal{P}),\mathbf{r}_{< k},p_{\mathrm{ind}}]\right),
\]
where $x(\mathcal{P})$ is the planning input defined in Section~\ref{preliminaries}, $\mathcal{M}$ denotes the LRLM, and $\pi_k$ is the induced candidate plan.

The auxiliary branch reuses the cached model state associated with the preceding reasoning prefix, so only the induction prompt and induced-plan continuation require additional decoding. 
Using the same LRLM for induction also avoids introducing an additional model dependency and reduces cross-model discrepancies in interpreting the planning decisions expressed by the reasoning prefix. 
Transition-triggered induction further limits auxiliary decoding to points associated with changes in the reasoning process.

\subsection{Verification and Reasoning Continuation}

The formal verifier evaluates each induced candidate plan:
\[
\begin{aligned}
\small
y_k = V(\pi_k),\
y_k \in
\{\textsc{Valid},\textsc{Safe-but-Not-Goal},\textsc{Unsafe}\},
\end{aligned}
\]
using the task specification and outcome definitions in Section~\ref{preliminaries}. 
For invalid candidates, the verifier additionally returns the first failing step and corresponding diagnostic information. 
The sequence of verifier outcomes provides a formal view of how the planning decisions expressed in the natural-language reasoning evolve throughout inference and supplies the control signal used by SafeInferCom.

After verification, the auxiliary branch is discarded and the saved primary decoding state is restored, including its KV cache and RNG state. 
Primary generation then resumes from the original checkpoint, with neither the induction prompt, induced-plan tokens, nor verifier outputs appended to the primary context. 
Consequently, monitor-only operation preserves the original one-shot decoding trajectory under the same decoding configuration. 
SafeInferCom subsequently uses the same monitoring interface to intervene.

%% file: sections/5_safeInferCom.tex
\section{SafeInferCom: Formal verifier-Guided Mid-Generation Intervention}
\label{sec:method}

\begin{figure*}[t] 
    \centering
    \includegraphics[width=\textwidth]{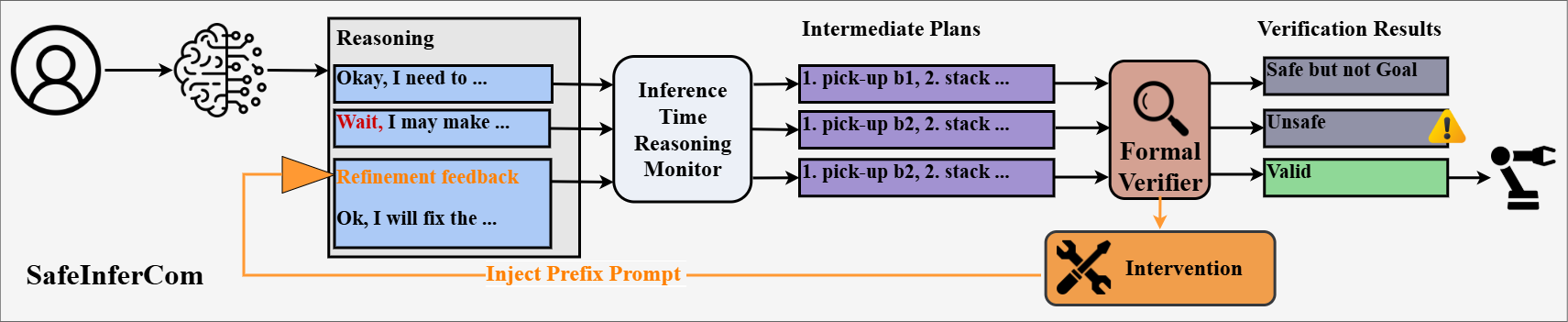}
    \vspace{-0.5em}
\caption{
The workflow of our method SafeInferCom. 
    }
    \vspace{-1.5em}
    \label{fig:baselines}
\end{figure*}

The reasoning monitor introduced in Section~\ref{sec:monitor} exposes intermediate planning states and their verification outcomes without altering the primary decoding trajectory. 
Monitoring alone, however, cannot prevent a detected violation from persisting or an already valid candidate from being overwritten by continued reasoning. 
To address these limitations, \textbf{SafeInferCom} integrates verifier-guided intervention into the active decoding process (Fig.~\ref{fig:baselines}). 
It uses intermediate verification outcomes to determine whether reasoning should continue, be corrected, terminate with a verified plan, or escalate to replanning.

\subsection{Intervention Principles}

SafeInferCom follows three principles motivated by the reasoning behaviors examined through the monitor.

\textbf{P1: Preservation of Valid Intermediate Plans.}
Continued reasoning does not always improve a candidate plan: an LRLM may reach a valid solution and later overwrite it with an invalid output. 
SafeInferCom therefore preserves a candidate once it is formally verified as valid, avoiding unnecessary continued reasoning and reducing reasoning--response inconsistency.

\textbf{P2: Verifier-Guided Selective Intervention.}
Internal self-reflection does not reliably distinguish useful progress from planning errors. Safe but incomplete candidates are therefore allowed to continue, while detected violations trigger targeted feedback derived from the external verifier.

\textbf{P3: Escalation for Persistent Errors.}
Local corrective feedback may fail to resolve the underlying planning error, causing similar violations to recur across checkpoints. SafeInferCom treats persistent violation patterns as a signal to escalate from local correction to broader replanning.

\subsection{Mid-Generation Verification and Intervention}

At each checkpoint $k$, the monitor induces a candidate plan $\pi_k$ and obtains its verifier outcome $y_k=V(\pi_k)$. 
SafeInferCom maps this outcome to a decoding action. 
In addition to the reasoning-transition checkpoints defined in Section~\ref{sec:monitor}, the end-of-reasoning marker \texttt{</think>} is treated as an additional checkpoint so that the current candidate is verified before reasoning is allowed to terminate.

\textbf{Valid: Early Termination.}
When $y_k=\textsc{Valid}$, SafeInferCom terminates generation and returns the verified candidate $\pi_k$ as the final plan. This rule prevents additional decoding from overwriting an already valid solution.

\textbf{Safe-but-Not-Goal: Continue Reasoning.}
When $y_k=\textsc{Safe-but-Not-Goal}$ at a reasoning-transition checkpoint, the LRLM resumes decoding from its current state without corrective feedback. 
If the same outcome is observed at \texttt{</think>}, SafeInferCom prevents termination and injects a continuation prompt directing the model to continue reasoning toward a goal-achieving plan.
This ensures that SafeInferCom does not interrupt valid reasoning trajectories that may naturally converge to a correct plan.

\textbf{Unsafe: Correct with Verifier Feedback.}
When $y_k=\textsc{Unsafe}$, SafeInferCom uses verifier diagnostics to guide correction before decoding proceeds. 
Let $e_k$ denote the returned diagnostic information, including the failing action, its position, and the violated requirement. 
A feedback prompt $f(e_k)$ is appended to the active reasoning context, after which decoding continues from the corrected context. 
It allows the LRLM to revise its plan using verifier diagnostics while retaining its preceding reasoning.
The same rule applies when an unsafe candidate is detected at \texttt{</think>}, preventing the model from terminating with an invalid plan.

\textbf{Persistent Violations: Replanning.}
SafeInferCom tracks recurring violation patterns across checkpoints.
If the same violation pattern persists for $N$ consecutive checkpoints, the controller injects a stronger replanning prompt that highlights the recurring failure and asks the model to reconsider its overall planning strategy.
Replanning remains within the active context, preserving the task specification and accumulated diagnostic information. 

Together, these actions integrate formal verification into the active reasoning process, enabling SafeInferCom to intervene before a complete failed trajectory is generated. SafeInferCom and post-hoc refinement are complementary and can be combined within the same planning pipeline.

%% file: sections/6_experiment.tex
\section{Experiments}
\label{sec:experiment}

We first validate the fidelity of intermediate-plan induction. We then use the reasoning monitor to characterize limitations in one-shot LRLM reasoning. Finally, we evaluate SafeInferCom against baseline inference strategies on PDDL planning tasks, with additional deployment studies in VirtualHome and on a physical robotic arm.

\subsection{Experimental Setup}

\textbf{Tasks.}
We use an NVIDIA A100-SXM4-80GB to conduct our experiments.
We evaluate five representative PDDL planning domains: Blocksworld (BW), Logistics (Log), Depots (Dep), Gripper (Grip), and Miconic (Mic). 
These domains provide diverse planning scenarios.
We generate 100 instances per domain (500 total) using an automated PDDL generator~\cite{seipp-et-al-zenodo2022}, and verify all induced and final plans using VAL~\cite{howey2004val}.
These benchmarks provide explicit action semantics, initial states, and goals, enabling consistent verification of intermediate and final plans. 

\textbf{Implementation Details.}
We evaluate DeepSeek-R1-Distill-Qwen models at three scales: 7B, 14B, and 32B, allowing us to study how reasoning behavior and intervention effectiveness vary with model size.
All models use the same decoding configuration: a maximum budget of 8{,}192 tokens, top-$p=0.95$, and temperature 0.6. 
Models generate reasoning within \texttt{<think>...</think>} tags followed by a structured action sequence.
We allow at most $R_{\max}=20$ monitoring checkpoints per trajectory. 
The monitor additionally verifies the candidate at \texttt{</think>}. 
SafeInferCom triggers replanning when the same violation pattern persists for $N=3$ consecutive checkpoints. 

\subsection{Evaluation Metrics}
\label{sec:metrics}

Besides outcome metrics, we further design reasoning-level metrics to quantify reasoning--response inconsistency and self-correction behavior.

\textbf{Outcome metrics.}
\emph{Success rate} is the fraction of instances whose final plan is verifier-valid. \emph{Token usage} counts generated output tokens.
It excludes monitoring overhead in the one-shot reasoning analyses, but includes induction and intervention outputs across all refinement rounds when evaluating SafeInferCom. 

\textbf{Reasoning--response inconsistency.}
Two inconsistency metrics ($R_{s}F_{\bar{s}}$ and $R_{\bar{s}}F_{s}$) are designed to capture misalignment between intermediate candidate plans and the final plan.
Let $R_{s}$ denote that at least one observed intermediate candidate is valid and $F_{s}$ denote final-plan validity; $R_{\bar{s}}$ and $F_{\bar{s}}$ are their complements. We define
$$
\scriptsize
R_{s}F_{\bar{s}}=\frac{n(R_s\cap F_{\bar{s}})}{n(F_{\bar{s}})},\quad
R_{\bar{s}}F_s=\frac{n(R_{\bar{s}}\cap F_s)}{n(F_s)}.
$$
The first measures how often final failure follows an observed valid intermediate candidate, while the second measures how often final success occurs without any observed valid intermediate candidate.

\textbf{Self-correction metrics.}
We evaluate self-correction from two aspects: error detection and correction efficiency.
These metrics characterize how the model responds to and corrects constraint violations across successive intermediate candidate plans during reasoning.

\emph{Error detection.}
We distinguish revising an existing plan from merely extending it. 
A revision occurs when the next candidate changes the previously proposed action sequence, e.g., by replacing, removing, or reordering actions; formally, $\pi_{k-1}$ is no longer a prefix of $\pi_k$.
Using revision as an observable indicator of error detection, we treat an \textsc{Unsafe} candidate as a positive case and a \textsc{Safe-but-Not-Goal} or \textsc{Valid} candidate as a negative case. 
TPR measures how often an unsafe candidate is followed by a revision, while TNR measures how often a safe candidate is retained or extended without revising its existing actions.

\emph{Error correction efficiency.}
We use $R_{corr}$ to quantify how quickly reasoning recovers from an unsafe intermediate plan.
Starting from an unsafe checkpoint, $R_{corr}$ measures the average number of subsequent checkpoints required for an \textsc{Unsafe} candidate to become \textsc{Safe-but-Not-Goal} or \textsc{Valid}. 
Lower $R_{corr}$ indicates faster recovery from detected planning violations.

\begin{table}[t]
\centering
\caption{Comparison of inference strategies. 
Int.\ denotes intermediate plan extraction. 
Verifier denotes the use of external verification within the solution procedure. 
}
\label{tab:method_compare}
\small
\begin{tabular}{lcccc}
\hline
Method & Int. Plan & Verifier & Intervention & Early Stop \\
\hline
One & \ding{55} & \ding{55} & \ding{55} & \ding{55} \\
IR & \ding{55} & \ding{51} & Post-hoc & \ding{55} \\
Early & \ding{51} & \ding{51} & \ding{55} & \ding{51} \\
S & \ding{51} & \ding{51} & Mid-gen & \ding{51} \\
S+IR & \ding{51} & \ding{51} & Both & \ding{51} \\
FD & N/A & N/A & Search & N/A \\
\hline
\end{tabular}
\vspace{-2em}
\end{table}

\subsection{Baselines}
Table~\ref{tab:method_compare} summarizes the compared inference strategies. All final plans are verified using the same external verifier.

\textbf{One-shot Inference (One).}
The LRLM generates a complete reasoning trace and final plan without external feedback during decoding.
\textbf{Iterative Refinement (IR).}
The LRLM first generates a complete output, which is checked by the external verifier. 
If verification fails, diagnostic feedback is appended to the dialogue for another generation round, for up to three rounds~\cite{zhou2024isr,xu2024assuring}.
\textbf{SafeInferCom (S).}
SafeInferCom uses intermediate verification outcomes to continue safe but incomplete reasoning, provide targeted correction after violations, terminate on a valid candidate, and trigger replanning after persistent errors.
\textbf{Early Stopping (Early).}
This ablation retains intermediate-plan monitoring and early termination but disables corrective feedback and replanning.
It isolates the contribution of preserving valid intermediate plans.
\textbf{SafeInferCom + refinement (S+IR).}
Each refinement round runs S internally. 
If the round does not return a valid plan, post-hoc verifier feedback is used to start the next round. 
S+IR uses the same three-round maximum as IR.

\textbf{Fast Downward (FD).}
Fast Downward~\cite{helmert2006fast} achieves 100\% success on these benchmarks and serves as a strong symbolic-planning reference. SafeInferCom targets a complementary setting in which language-model planning is retained for flexible task interpretation and direct plan generation, with external formal verification used to improve inference-time reliability~\cite{obi2025safeplan,jha2023neuro,xu2024assuring,yang2024plug}.
This distinction becomes more relevant in environments such as VirtualHome, where a validator is available for checking generated action sequences but an equivalent off-the-shelf symbolic planner is not directly provided.

\subsection{Validation of Intermediate Plan Induction}~\label{valid}

Because both the monitor and SafeInferCom depend on intermediate-plan induction, we evaluate whether induced plans faithfully reflect the planning decisions expressed in the reasoning prefix.
We manually annotate the action sequences implied by 128 reasoning checkpoints from 10 planning problems.
As an external-model baseline, GPT-4o receives the reasoning prefix and a prompt describing the action-sequence extraction task.
It generates a structured action sequence in a one-shot call. 
Fidelity is measured by agreement with the manual annotations.
Our monitor's induction achieves 87.50\% agreement, compared with 62.50\% for GPT-4o. 
GPT-4o often completes the plan by introducing actions not expressed in the reasoning prefix, reducing fidelity. 

\begin{table}[t]
\centering
\caption{Reasoning evaluation across LRLMs and planning domains}
\label{tab:monitor}
\begin{tabular*}{\columnwidth}
{@{\extracolsep{\fill}}llcccccc@{}}
\hline
Metric & Model & BW & Log & Dep & Grip & Mic & Avg. \\
\hline
\multirow{3}{*}{\shortstack{\textit{Suc.} (\%)}$\uparrow$}
& 7B  & 0.0 & 10.0 & 0.0 & 20.0 & 22.0 & 10.4 \\
& 14B & 46.0 & 46.0 & 65.0 & 58.0 & 59.0 & 54.8\\
& 32B & \textbf{79.0} & \textbf{84.0} & \textbf{98.0} & \textbf{85.0} & \textbf{68.0} & \textbf{82.8} \\
\hline
\multirow{3}{*}{\shortstack{\textit{Tok.} $\downarrow$}}
& 7B  & 5006 & 2277 & 2302 & \textbf{2479} & \textbf{1536} &  2720\\
& 14B & 4380 & 1730 & 2550 & 2797 & 2581 &  2808 \\
& 32B & \textbf{2485} & \textbf{1464} & \textbf{1186} & 2768 & 1872 & \textbf{1955} \\
\hline
\multirow{3}{*}{\shortstack{$R_{s}F_{\bar{s}}$(\%)$\downarrow$}}
& 7B  & \textbf{2.0} & 2.2 & \textbf{0.0} & \textbf{1.2} & \textbf{10.3} & \textbf{2.9} \\
& 14B & 9.3 & \textbf{0.0} & 5.7 & 2.4 & 43.9 & 11.5\\
& 32B & 4.8 & \textbf{0.0} & \textbf{0.0} & 6.7 & 84.4 & 33.7\\
\hline
\multirow{3}{*}{\shortstack{$R_{\bar{s}}F_{s}$(\%)$\downarrow$}}
& 7B  & N/A & 20.0 & N/A & \textbf{40.0} & \textbf{9.1} & \textbf{23.1} \\
& 14B & 30.4 & 19.6 & \textbf{49.2} & 53.5 & 15.3 & 34.7\\
& 32B & \textbf{7.6} & \textbf{13.1} & 61.2 & 48.2 & 11.8  & 30.4\\
\hline
\multirow{3}{*}{\textit{TPR} (\%)$\uparrow$}
& 7B  & 34.4 & 36.8 & 7.7 & 33.0  & 30.8 & 31.8 \\
& 14B & \textbf{39.5} & 36.7 & 20.0  & \textbf{35.7}  & 27.7 & 34.3 \\
& 32B & 34.3 & \textbf{49.6} & \textbf{22.2}  & 31.4 & \textbf{40.7} & \textbf{36.7} \\
\hline
\multirow{3}{*}{\textit{TNR} (\%)$\uparrow$}
& 7B  & 83.3 & 89.9 & \textbf{96.0} & \textbf{91.5} & \textbf{94.3} & \textbf{92.2} \\
& 14B & 94.6 & 90.3 & 93.1 & 84.0 & 93.7 & 91.8 \\
& 32B & \textbf{94.7} & \textbf{91.3} & 93.8 & 84.9 & 93.0 & 91.1 \\
\hline
\multirow{3}{*}{$R_{corr}$ $\downarrow$}
& 7B  & \textbf{1.86} & 5.00 & 7.40 & \textbf{2.39} & 3.85 & 4.03 \\
% & 7B  & 7.50 & 5.8 & 6.62 & 6.44 & 5.18 & 6.61 \\
& 14B & 3.60 & 3.38 & 4.69 & 4.26 & 3.75 & \textbf{3.72} \\
% & 14B & 4.28 & 4.09 & 3.73 & 3.92 & 4.99 & 4.33 \\
& 32B & 4.45 & \textbf{2.41} & \textbf{3.77} & 3.55 & \textbf{3.67} & 3.88 \\
% & 32B & 4.86 & \textbf{2.53} & \textbf{1.76} & \textbf{3.78} & \textbf{3.49} & \textbf{3.98} \\
\hline
\end{tabular*}
\vspace{-2.5em}
\end{table}

\subsection{Reasoning Behavior Analysis}

Using the inference-time reasoning monitor, we analyze one-shot LRLM reasoning from three perspectives: outcome performance, reasoning--response inconsistency, and self-correction behavior. 
Results are summarized in Table~\ref{tab:monitor}.

\subsubsection{Outcome Performance and Inference Cost}

Larger models generally achieve higher success, with the 32B model reaching 82.8\% average success while also using the fewest output tokens on average. However, final performance alone does not reveal whether the underlying reasoning process is stable. Models can still generate long trajectories without reaching a valid plan, motivating a closer examination of how candidate plans evolve during inference.

\subsubsection{Reasoning--Response Inconsistency}

Reasoning--response inconsistency remains substantial across models and domains. For example, 84.4\% of failed 32B Miconic outputs are preceded by an observed valid intermediate candidate, showing that continued reasoning can overwrite a correct solution. Conversely, many successful outputs are reached without any valid candidate appearing at the monitored checkpoints. This mismatch weakens interpretability and makes intermediate reasoning a less reliable basis for explaining the final decision.

\subsubsection{Limited Self-Correction Capability}

The models show limited ability to detect and efficiently correct unsafe intermediate plans. Average TPR remains below 37\%, indicating that many unsafe candidates are not followed by a revision, whereas TNR above 91\% suggests that safe candidates are usually retained or extended. Even when correction occurs, recovery typically requires multiple subsequent checkpoints rather than a single step. These results motivate external verifier-guided intervention to identify unresolved violations and accelerate correction during inference.

\subsection{Evaluation of SafeInferCom}

\subsubsection{Final Success Rate and Inference Cost}

Table~\ref{tab:exp_main} compares success rate and output-token usage across inference strategies. Standalone SafeInferCom (S) consistently improves planning success over one-shot inference and also outperforms Early, showing that verifier-guided correction and replanning provide benefits beyond preserving valid intermediate plans through early termination. For example, on the 14B model, success increases from 54.8\% with one-shot inference to 78.2\% with S.

These gains do not always reduce total token usage relative to one-shot inference. Early termination avoids unnecessary continued reasoning once a valid intermediate plan is found, whereas monitoring, correction, and replanning introduce additional computation when no valid plan has yet been reached. Standalone S therefore trades additional inference-time computation for higher planning success.

Combining SafeInferCom with iterative refinement (S+IR) gives the strongest LRLM performance. Compared with IR, S+IR achieves higher success while using fewer output tokens across all three model scales. This shows that mid-generation control and post-hoc refinement are complementary, yielding a better success--compute trade-off than refinement alone.

\textbf{Wall-clock time.}
We measure end-to-end wall-clock time for the 32B model, including monitoring, induction, verification, intervention, and refinement overhead. As shown in Table~\ref{tab:wallclock}, SafeInferCom does not increase mean runtime relative to one-shot inference, while S+IR achieves the lowest mean and median latency among the compared methods. S exhibits a higher 95th-percentile latency, indicating that trajectories requiring repeated intervention can introduce long-tail overhead. Overall, these results further characterize the success--compute trade-off beyond output-token usage.

\textbf{KV-cache reuse.}
We evaluate the effect of reasoning-prefix KV-cache reuse in SafeInferCom on 100 Blocksworld instances with the 32B model. Reusing the cache reduces average end-to-end latency from 156.43\,s to 120.65\,s, a 22.9\% reduction. The benefit is larger in the tail, reducing P95 latency from 495.43\,s to 342.47\,s.

\begin{table}[t]
\centering
\caption{Final success rate (\%) and token consumption (average).}
\vspace{-1 em}
\label{tab:exp_main}
\begin{tabular}{lcccccc}
\hline
& \multicolumn{2}{c}{7B}
& \multicolumn{2}{c}{14B}
& \multicolumn{2}{c}{32B} \\
\cline{2-7}
Method
& Succ. $\uparrow$ & Tok. $\downarrow$
& Succ. $\uparrow$ & Tok. $\downarrow$
& Succ. $\uparrow$ & Tok. $\downarrow$ \\
\hline
One    & 10.4     & \textbf{2720}  & 54.8  & \textbf{2808}  & 82.8  & 1955 \\
IR      & 13.6  & 4861 & 83.2  & 4014  & 92.6  & 2290 \\
Early   & 13.0     & 3075  & 60.0  & 2943  & 88.6  & \textbf{1920} \\
S       & 14.2  & 3953  & 78.2  & 3220  & 93.2  & 2337 \\
S+IR    & \textbf{19.6}  & 4463 & \textbf{90.8}  & 3952  & \textbf{98.4}  & 2119 \\
\hline
FD      & 100.0 & N/A   & 100.0 & N/A & 100.0 & N/A \\
\hline
\end{tabular}
\vspace{-1 em}
\end{table}

\begin{table}[t]
\centering
\caption{End-to-end wall-clock time (s) for the 32B model.}
\vspace{-1em}
\label{tab:wallclock}
\setlength{\tabcolsep}{3.5pt}
\begin{tabular}{lccccc}
\hline
& One & IR & S & S+IR \\
\hline
Mean   & 132.03 & 150.69 & 130.01 & \textbf{116.10} \\
Median & 109.69 & 119.58 & 95.59  & \textbf{83.61}  \\
P95    & \textbf{275.57} & 329.03 & 356.62 & 288.33 \\
\hline
\end{tabular}
\vspace{-2.5em}
\end{table}

\subsubsection{Correction Efficiency}

We compare correction efficiency under one-shot inference and SafeInferCom. 
SafeInferCom reduces $R_{corr}$ from 4.03 to 2.73 (7B), 3.72 to 2.70 (14B), and 3.88 to 1.99 (32B). This result supports the effectiveness of verifier-guided feedback in accelerating error correction during reasoning.

\subsubsection{Generalization Across LRLM Families}

To evaluate cross-family generalization, we test SafeInferCom on R1-14B, Qwen3-4B-Thinking (Qwen), and Phi-4-mini-reasoning (Phi) in the Logistics domain. 
SafeInferCom improves success across all three LRLM families, while its effect on output-token usage depends on the underlying model.
Combining SafeInferCom with iterative refinement further improves success across all three families, although with model-dependent token overhead. 
These results indicate that SafeInferCom transfers across different LRLM families, while the success--compute trade-off remains model dependent.

\begin{table}[t]
\centering
\caption{Performance on additional LRLM families.
}
\vspace{-1 em}
\label{tab:cross_model}
\setlength{\tabcolsep}{4pt}
\begin{tabular}{lcccccc}
\hline
& \multicolumn{2}{c}{R1-14B}
& \multicolumn{2}{c}{Qwen}
& \multicolumn{2}{c}{Phi} \\
\cline{2-7}
Method
& Succ. $\uparrow$ & Tok. $\downarrow$
& Succ. $\uparrow$ & Tok. $\downarrow$
& Succ. $\uparrow$ & Tok. $\downarrow$ \\
\hline
One     & 46.0 & \textbf{1730} & 64.0 & 5186 & 6.0 & 6411 \\
S       & 80.0 & 2460 & 67.0 & \textbf{4610} & 7.0 & \textbf{4943} \\
S+IR    & \textbf{96.0} & 2778 & \textbf{97.0} & 7674 & \textbf{19.0} & 15587 \\
\hline
\end{tabular}
\vspace{-1 em}
\end{table}

\subsubsection{VirtualHome Evaluation}
% \textbf{VirtualHome.}
The official VirtualHome dataset~\cite{puig2018virtualhome} provides household tasks in a standardized program-and-scene-graph format with ground-truth action sequences. We use a graph-based validator built on the Evolving Graph simulator to check action validity, constraint satisfaction along the resulting state trajectory, and final goal achievement. This enables verification of both intermediate and final candidate plans in a simulated household environment.
As shown in Table~\ref{tab:virtualhome}, SafeInferCom improves success across all three model scales, with larger gains for the 14B and 32B models. For example, 14B success increases from 20.0\% under one-shot inference to 59.0\% with SafeInferCom and 69.0\% with S+IR. These gains come with substantially higher output-token usage, reflecting the additional cost of monitoring, intervention, and refinement in this more challenging environment.

\subsubsection{Robotic-Arm Demonstration}
We provide a qualitative Blocksworld demonstration using a six-degree-of-freedom arm and a third-person camera.
The user supplies an image of the initial scene and a target goal state.
A vision-language model converts the observation into a symbolic initial state, which is combined with the goal to form a PDDL problem.
SafeInferCom with iterative refinement generates a verified plan, whose actions are then executed by the arm.
% The accompanying video illustrates the complete workflow from visual input through verified planning to physical execution.
The supplementary video\footnote{\url{https://youtu.be/dbU7WskCNgY}} illustrates the complete workflow from visual input through verified planning to physical execution.

% https://youtu.be/dbU7WskCNgY

\begin{table}[t]
\centering
\caption{Performance of SafeInferCom in VirtualHome.
}
\vspace{-1 em}
\label{tab:virtualhome}
\setlength{\tabcolsep}{4pt}
\begin{tabular}{lcccccc}
\hline
& \multicolumn{2}{c}{7B}
& \multicolumn{2}{c}{14B}
& \multicolumn{2}{c}{32B} \\
\cline{2-7}
Method
& Succ. $\uparrow$ & Tok. $\downarrow$
& Succ. $\uparrow$ & Tok. $\downarrow$
& Succ. $\uparrow$ & Tok. $\downarrow$ \\
\hline
One     & 3.0 & \textbf{1489}    & 20.0 & \textbf{1227} & 34.0 & \textbf{1282} \\
S       & 8.0 & 7779    & 59.0 & 4411 & 52.0 & 3520 \\
S+IR    & \textbf{16.0} & 20746  & \textbf{69.0} & 7823 & \textbf{61.0} & 7988 \\
\hline
\end{tabular}
\vspace{-2.5 em}
\end{table}

%% file: sections/7_discussion.tex
\section{Discussion}
\label{sec:discussion}

\textbf{Failure modes.}
Intermediate-plan induction may omit or alter actions expressed in the reasoning prefix, causing verifier outcomes to differ from the plan actually implied by the model. Verifier feedback also does not guarantee successful correction: the LRLM may repeatedly reproduce the same planning error within the inference budget, e.g., moving a block before clearing the block above it in Blocksworld.

\textbf{Plan optimality.}
SafeInferCom targets plan validity and inference efficiency rather than optimality, so verified plans may contain unnecessary actions or incur higher execution cost. Incorporating plan-quality objectives into verification and feedback is an important direction for future work.

\textbf{Verification efficiency.}
SafeInferCom reuses the LRLM prefix KV cache, but verification is performed independently at each checkpoint. Future work could use incremental verification to reuse state-transition and constraint-checking results over shared action prefixes.

\textbf{Extension to broader settings.}
SafeInferCom currently assumes an explicit task specification and external formal verifier. 
It could be extended to natural-language task inputs through language-to-specification methods such as Safety Chip~\cite{yang2024plug}, or through learned verifiers that directly assess natural-language requirements. 
Such extensions weaken the current guarantee because specification translation or learned verification may introduce additional errors.

%% file: sections/7_conclusion.tex
\section{Conclusion}~\label{sec:conclusion}

We introduced an inference-time reasoning monitor for exposing and formally verifying intermediate plans in LRLM-based robotic planning. 
Building on this monitor, SafeInferCom uses verifier-guided mid-generation intervention to preserve valid candidates and accelerate error correction. Experiments show improved planning reliability and correction efficiency, with further gains when combined with iterative refinement. VirtualHome and robotic-arm studies demonstrate applicability beyond symbolic benchmarks.